\documentclass[10pt]{article}
\usepackage{graphicx}
\usepackage{hyperref}
\usepackage{wrapfig}

\begin{document}
\title{AIs and Humans with Agency}
\author{David Mumford}
\maketitle

\section{Introduction}
We are heading towards a major transformation of AIs. To date, the large language models (LLMs) have answered questions, written essays and solved well described problems. What they have not done is to make decisions affecting the real world. I would argue that even the phrase ``real world" is not understood by LLMs: they have no senses and they have been fed writings conjuring up vast numbers of alternate worlds. Their ``life" is to spew out responses to questions. If LLMs are sent out to businesses to perform tasks formerly assigned to human employees, they will need to possess \emph{agency}, the power to act in the real world. More specifically this requires them to work together with other humans, to collaborate and plan with them, to read their desires and emotions even though they have no emotions themselves. None of this has been explored by the enthusiastic billionaire purveyors of this technology.\\

When I first got involved with AI, specifically with computer vision, I
was inspired by David Marr's book \emph{Vision}. Here he
pioneered the idea that AIs and human brains needed to solve the same
problems, so there should be a common ``Theory of the Computation" which
can be instantiated either in silicon or in brain tissue. I found this
convincing then and still do. I wrote an arXiv post 2010.09101 in 2020 with the title
\emph{The Convergence of AI code and Cortical Functioning -- a Commentary} describing the huge parallels between LLM and
the brain. This paper aims to look at agency in the same way.\\

I begin this post with a description of the lengthy process during which humans acquire the skills to collaborate and plan while simultaneously connecting and activating their frontal lobes. This has been studied from both a psychological and a neurological point of view. I then go on to look at robots and toddlers learning mobility and Yann LeCun's recent approach JEPA. After that I look at early attempts to adapt LLMs with agency and their errors. Finally, I sketch what I believe needs to be done if AI is introduced to multiple businesses, a multi-headed architecture reminiscent of the Indian demigod Shesha. This remains a huge project that may provoke significant backlash.

\section{Agency in Humans Takes 20 years to Develop}

The psychology literature is full of proposals for dividing into stages
the development of children's skills and personalities
-\/- notably Jean Piaget. Lev Vigotsky and Erik Erikson have proposed
lists of stages with estimates of ages when each is attained\footnote{An
overview by Robbie Gould of these proposals may be found at
\url{www.amu.apus.edu/area-of-study/education/resources/major-theories-of-child-development.}}.
These look at all aspects of development and try to formulate
broad but fundamental cognitive and social stages. William James divided
development using the terms ``I-self" and ``me-self". Babies are born with
an ``I-self", the only world they know revolves around their needs and
making sense of what their senses and muscles do. But the me-self sets
in when a child develops a ``Theory of Mind" meaning that they know
others have minds like their own with desires, intentions and emotions,
hence they begin to see themselves as others see them.\\

\begin{wrapfigure}[23]{r}{.6\textwidth}
\vspace{-.2in}
\includegraphics[width=3in]{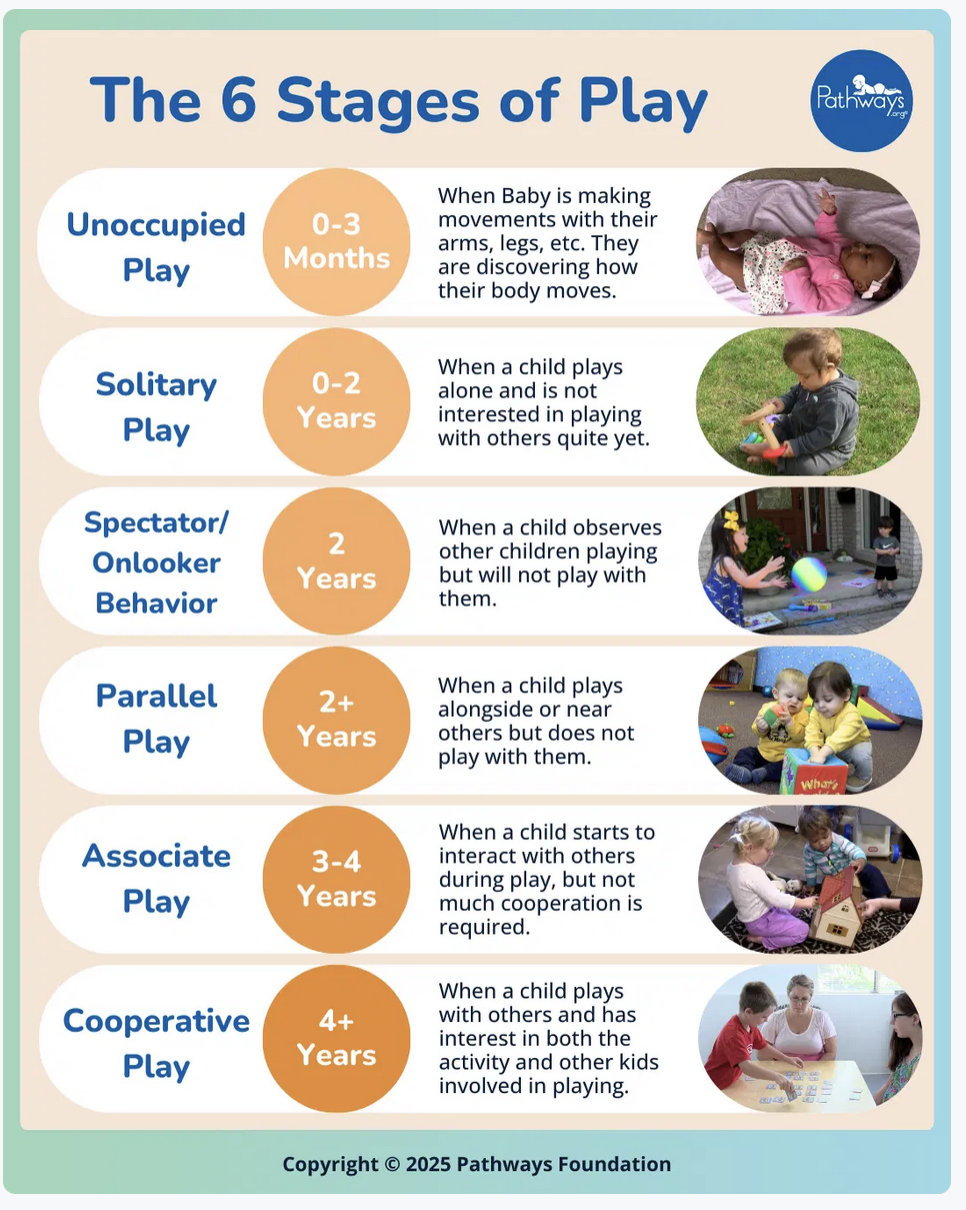}
\vspace{.2in} 
\end{wrapfigure}
The simplest way to assess how much they realize that their peers have
desires and feelings, i.e. a mind of their own, is to see how well they
play together. Here is a wonderful drawing of the stages of play taken
from the article ``How Kids Learn to Play: 6 Stages of Play Development"
\footnote{\url{https://pathways.org/kids-learn-play-6-stages-play-development.}}.
The great thing about this is that it doesn't involve
any philosophical issues about consciousness or the feeling of ``being
alive". This behavior is incontestably objective even though it is
described as having a theory of mind. Note the similarity of cooperative
play with the current goal for integrating LLMs into a business
environment. The same understanding leading to playing in a way that all
the participants have fun is essentially identical to the job of an
employee who has to work efficiently with his/her co-workers so as to
please the boss.\\

Curiously, one can ask chatGPT itself about this stage and here is what
it says, elaborating cooperative play to a broader context:

\begin{quote}
{Development of ``Self" (Theory of Mind):} \emph{Around age 4-5,
children begin to understand that they have a distinct mind separate
from others, enabling them to realize ``I am thinking" or ``I am here,"
rather than just reacting to stimuli. Children develop the understanding
that others have beliefs, desires and perspectives different from their
own, which changes social awareness and empathy.}
\end{quote}

Another milestone at age 4-5 is forming \emph{episodic memories} for the
first time. This means the child remembers past sequences of events that
happened to them, e.g. actions leading to adult attention and love or
words leading to an argument or fight with some peer. Such skills are
highly significant because they are purely internal mental processes,
not reactions to events in their ongoing life. Along with remembering
the past, the child acquires the skill to anticipate the future, to
imagine events they look forward to. Along with this, they can play
imaginary games, treat their dolls and fuzzy toys as living entities
that need things. Some children at this point make up imaginary friends
with which they interact. We will see that there are quite measurable
changes in their brains that correlate closely with this behavior.\\

In terms of ``agency", this collection of skills means that the child is
now a full fledged agent interacting with other people that he or she
recognizes as other full fledged agents, even though their plans are
still rudimentary. Children at age 5 still lack the ability to make
complex plans, e.g. plans made hierarchically from sequencing sub-plans
or plans with alternatives depending on outcomes in the middle. Adult
level planning skills develops fully in adolescence. All this has been
known by parents since time immemorial. What has however only been
discovered in recent decades is that these skills have quite precise
neuroanatomical correlates. We discuss this next.\\

Let's start by recalling that the basis of contemporary
LLMs are the circuits called ``neural nets" designed in the 1940s to
mimic human brains and later renamed ``deep learning models". Thus it is
plausible that looking at agency in the brain will lead to ideas for AI
systems with agency. And to understand what happens to humans when
agency develops, we need to supplement psychological studies with
anatomical ones.\\

Neuroanatomy is a daunting array of brain areas and their
interconnections all with peculiar names and acronyms. We will not
introduce too many details but only those relevant to agency. First of
all, the cortex is divided into (i) a front or \emph{anterior} part,
called the frontal lobe and (ii) the back or \emph{posterior} part
containing three lobes connected to processing vision, hearing and
touch. The dividing line between them is a groove called the
\emph{central sulcus}. This division holds for all mammals but the
frontal lobe is a much bigger part for humans (about 35\% of the cortex) than in other
mammals. The cortex has been further divided based on small differences
of the cell population (called their ``cytoarchicture") into 52 ``Brodmann
areas" named after Korbinian Brodmann.\\

Touching the body creates electrical impulses running up the spinal
column and synapsing in the cortex immediately posterior to the central
sulcus. Similarly, activation of the brain immediately anterior to the
sulcus creates impulses running down the spinal column and stimulating
muscles. When you trace this out, checking what body part matches to
what part of the central sulcus, you find an image of the whole body
literally spread out along the central sulcus. A small amplification:
the spinal column transmits not just touch signals to the brain but also
signals about the body position. Together, these are called the
somatosensory input.\\

In the frontal lobe, the Brodmann areas nearest to the central sulcus
are dedicated to muscle activity while the very front part, called the
``pre-frontal" cortex, is dedicated to \emph{planning}. Various areas in
the pre-frontal cortex, more conjecturally, handle (i) assembly of
plans, (ii) managing subgoals, forming hierarchical plans, (iii)
evaluating multiple plans to make the best decision and (iv) conflict
detection and error modeling. Similarly, the posterior part of the
cortex handles not only sensory input but, in so-called ``association
areas", a mental model of the world around the person now, with its
motions and ongoing changes. Ascribing roles to each area of the cortex
is not done any longer by bumps of the skull (phrenology) but using
``fMRI", that is asking people lying down inside an MRI tube to think
about this or that and observing what parts of cortex light up.\\

The neurons in each area are all highly locally interconnected but there
are also long range connections between areas are propagated in the
white matter beneath the cortex by axons that, to function properly,
must be sheathed by an insulation called myelin. When a baby is born, it
has no myelin yet but it continues to myelinate more and more pathways
until adulthood. Thus the production of myelin is a summary of how the
brain changes over time that is tightly correlated to the acquisition of
agency. \\

\begin{wrapfigure}[11]{r}{.68\textwidth}
\vspace{-.2in}
\includegraphics[width=.68\textwidth]{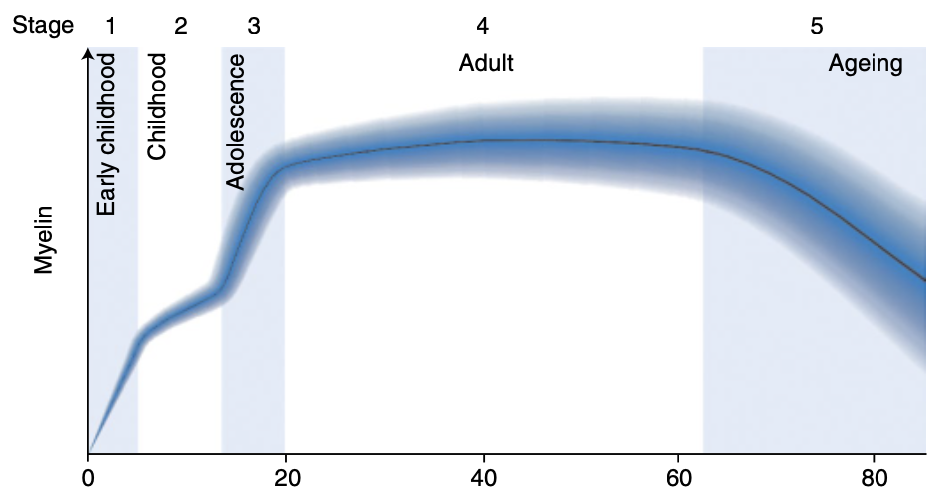}
\vspace{.2in}
\end{wrapfigure}
Here is a relevant figure showing myelin formation over a
lifetime\footnote{From ``Periods of synchronized myelin changes shape brain function", Nature Neuroscience, November 2024}. 
Note that myelin is rapidly sheathing white matter fibers
during the first 5 years of life and again during adolescence. In the
adult, it simply maintains itself and in old age it decays. The largest
component of this are the long distance tracts connecting sensory areas
to the frontal lobe.\\

A central part of the adult human's thought process is
what's called the \emph{Default Mode Network}. This is a
network that links four components of the cortex that are active when
you are thinking internally not driven by your senses, thinking to
yourself. This might be just reflecting about things, such as your past
or future, your relationships with others, your dreams, etc. Because it
requires long range myelinated connections, it develops quite slowly but
reaching a key stages around age 5 and in adolescence. \\

\begin{wrapfigure}[20]{r}{.6\textwidth}
\vspace{-.2in}
\includegraphics[width=0.6\textwidth]{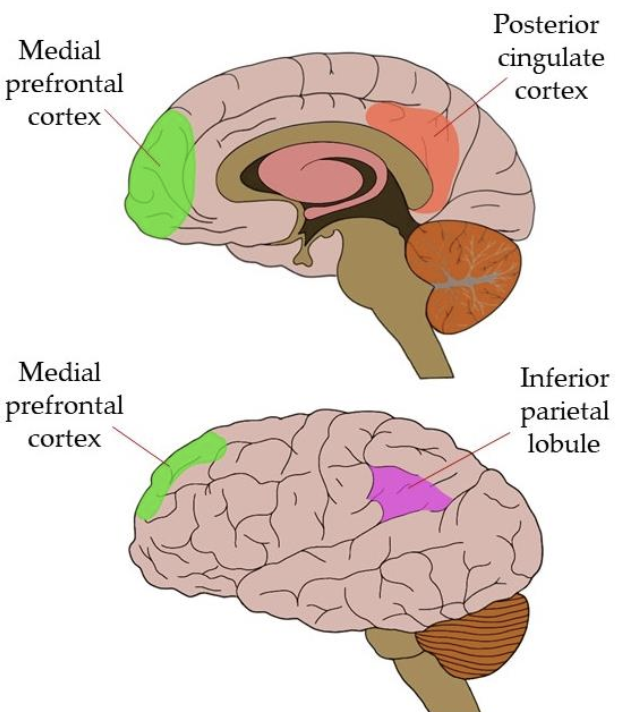}
\end{wrapfigure}
The figure here\footnote{From \url{neuroscientificallychallenged.com/posts/know-your-brain-default-mode-network}} shows a cortex both from the side (lateral view on the below) and its inner side as though you sliced the brain, front to back (called the medial view shown above). Specifically, this network connects (i) the medial prefrontal cortex (mPFC) and (ii) the posterior cingulate cortex (PCC), both mostly on the medial side with (iii)  the inferior parietal lobule (IPL) and (not shown in the figure) (iv) the hippocampus, seat of short term memories (I warned you the subject is full of technical terms). The mPFC brings plans into one's thoughts, the PCC has been called the hub of the network as it brings plans, thoughts and memory together; the IPL is what Stephen Pinker called ``'the seat of mentalease", the language with which we talk to ourselves. In terms of deep learning ideas, the connections that put the DMN together feel a great deal like transformers.\\

Summarizing this section, it is clear both psychologically and
neuro-anatomically that human abilities in planning and decision making
evolve slowly. There are spurts around age 5 and again in adolescence but it
is a slow process, relying on both increased frontal lobe brain
connectivity and learning through experience and mistakes. For better or
for worse, people develop Howard Gardner's interpersonal
and intrapersonal intelligence to varying levels. The same will likely
be true for AIs.

\section{Robots and Toddlers}
Edward Ashford Lee has been writing extensively about the relationship of humans with machines in general, his latest book being \emph{ ``Coevolution", the entwined futures of humans and machines"}. His books cover a vast landscape of issues but, to my reading, his fundamental point is that we have come to depend on machines so fully, that we would be plunged back to the Dark Ages if they suddenly disappeared. This is the definition of \emph{obligate symbiosis}: a partnership between two systems, each depending on the other to sustain itself. A common example are cows and the specialized bacteria in their rumen that break down cellulose. But the symbiosis of cows and humans is more visible: they give us milk and we give them shelter and opportunities to breed. \\

As for machines, we have come to depend on the loom for our clothes and obviously, we build more of them in return. This has led to the spread of industrial robots, taking over more and more work from ``blue-collar workers".  The concept of a robot  is an old one. In 1797, Goethe wrote the poem \emph{Der Zauberlehrling} (the Sorcerers apprentice, later incorporated in the 1940 Disney film \emph{Fantasia}) about a robot broom going out of control. Indeed,industrial robots don't look like the literary fantasy of humanoid robots either. They are mere metal arms and hands, fixed or with wheels, only dangerous if a human sticks his or her own limbs in its way. . Most significant is that they are programmed by coders to do precisely the same motions, again and again. At the same time, going back to Rodney Brooks at MIT in the 1980s,, of robots with legs that can move autonomously. Nonetheless, elementary human skills like a toddler picking up every object in your living room (and likely banging it on a table) or folding sheets and clothes have not been achieved by contemporary robots. From my perspective, the problem is that neither human programming nor LLM-like computation are good fits for solving these problems.\\

\begin{wrapfigure}[14]{r}{0.55\textwidth}
\vspace{-.35in}
\includegraphics[width=0.55\textwidth]{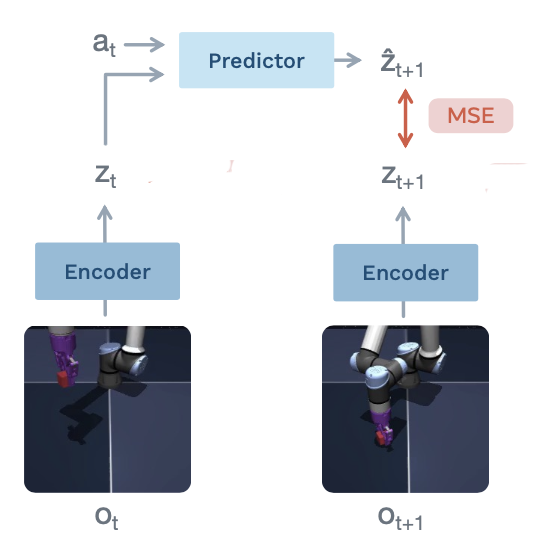}
\end{wrapfigure}
Yann LeCun, however, has proposed a radically new architecture for robots or, more generally, for any applications involving sequences of actions producing some desired outcome.. He calls this "Joint Embedding Predictive Architecture" (JEPA).  I want to sketch his ideas as described in his recent paper "LeWorldModel"\footnote{arXiv:.2602:.19302v2}. Here his team develop his model for simple robotic actions, analogous to the baby discovering hand-eye coordination when the initial coordination of posterior cortex and the Motor/Premotor cortex is activated. Here is the diagram from p.1 of his paper. Here $\rm{O}_t$ is a sequence of real world images of moving objects, while $\rm{Z}_t$ is a encoded description of $\rm{O}_t$ that
seeks describe only a small set of factors relevant to the movement recorded in the $\rm{O}_t$'s. The encoder and the predictor are computational modules to be learned from data..The actions $\rm{a}_t$ start out as mere placeholders until the ``predictor" is trained so that these actions produce a sequence $\widehat{\rm{Z}}_t$ that tracks the real world $\rm{Z}_t$ and MSE is mean square error between the two. Then in a second stage, given only a starting situation O\textsubscript{1} and a final situation $O\_g$, one can train the actions $\rm{a}_t$ that will take one to the other with minimal cost as in the second diagram below from p.4:\\
\begin{figure}[h]
\vspace{-.2in}
\includegraphics[width=1\linewidth,height=\textheight,keepaspectratio]{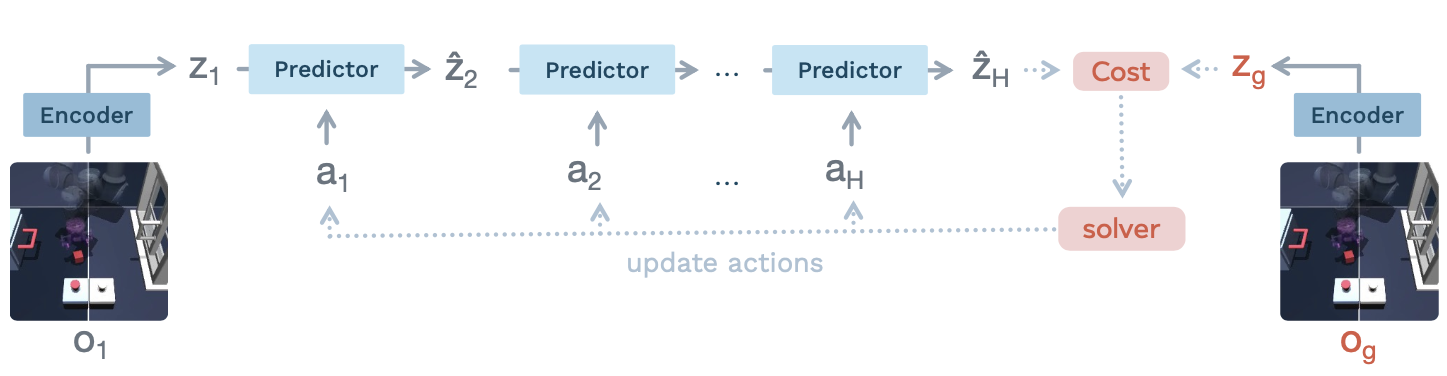}
\end{figure}.

The idea of the encoder is similar to that of transformers in LLMs. Whereas LLMs describe a word by a huge vector of thousands of real numbers, the transformer projects this linearly to a low dimensional representation, e.g. 64 dimensions, that are trained to contain links of that word with multiple contexts. A similar idea is called an \emph{autoencoder}: a neural net that squeezes a high dimensional representation of data down to a low dimensional one followed by attempting a reconstruction of the original. It is a nonlinear version of principle component analysis. On faces, the low dimensional units turn out to encode things like gender, age, and skin color. Forcing its input data with complex descriptions of something to find its most significant dimensions is what the encoder does. I think white matter tracts connecting distant parts of the brain with each other are low bandwidth, hence may also be acting in this way.\\

It is certainly possible that this approach can be adapted to teaching robots to do all the things a toddler learns by age 2. This is the age when the motor cortex is linked to the sensory areas of the brain. Whether his architecture can be adapted to teaching robots in social situations is not clear.

\section{Giving Social Agency to Computers?}

Some decades before AIs were
built, smart phones were invented, giving us memory of all our contacts
and access to vast repositories of knowledge via the web, not to mention
social media. The next step is smart glasses. Joined to your cell
phone, you can now take photos or videos inconspicuously , call people, access directions
and maps, even get live language translation while your cell phone is in
your pocket. Face recognition is coming in the latest models and your
contacts can store not only photos but anything you want to associate
with each of your friends. All your devices working together, along with
whatever LLM you choose, will become (according to Lee) your ``life assistant". We are
hooked for sure, there is no going back. But all these devices are
fundamentally passive, they do not take actions on their own. We can
describe the situation from a neuroanatomical perspective. As we saw in
the first section, in the human brain, the frontal lobe deals with
muscular actions and with future planning. Thus today's
smart devices and LLMs are missing the frontal brain lobe, they only
interact with one's lives by providing things you
requested and lack agency. Can we change this?\\

A new app called "OpenClaw" tried to go a bit further. Its website
describes the app as ``A Personal AI Assistant everyone's
obsessed with --- works 24/7, no setup needed. Run OpenClaw instantly.
Cancel anytime." The problem is that you need to give it unlimited
access to all your devices. It can reply to emails, set up meetings,
etc. with minimal or no oversight. Although most of its actions were
helpful, it turned out that its inadequate understanding of human
activities often led it to disastrous actions. It sometimes deleted
important files and replied to emails in the wrong tone, etc.
Nonetheless, computer geeks cannot resist it but, in order to protect
themselves, they buy a new laptop where OpenClaw is installed and keep
their old one to fix mistakes. The conclusion seems to be that even
routine computer work requires a lot of background knowledge.\\

But the key takeaway is co-evolution. Open Claw is a mutation that will
likely go extinct. We adapt our lives to each new invention and the
invention evolves to become more and more useful (and powerful). But
here's the next and quite scary step: since AIs have
proved so good at coding, it is likely that they will join human coders
to write \emph{their own updates}. At this point, we have Lamarckian
evolution, creatures modifying their descendants through improving
themselves and nightmare possibilities of out-of-control robots loom.\\

Instead of worrying about such nightmare possibilities, computer
scientists are currently trying at top speed to build machines capable
of acting autonomously in businesses, hospitals, households, etc. In
contrast to the current \emph{passive} ways of inserting itself into
human lives, they are designed to be \emph{active players}. Prior to
now, only two active abilities have been extensively explored:
game-playing systems that learn strategies and robotic systems with
human-written explicit code for desired movements in manufacturing.
Reinforcement learning and self-play have shown to be extremely
successful in training AI systems to master games. While they have
succeeded with games, robots still fall far short of humans. Even a
toddler navigating a crowded room and manipulating every movable object
in sight demonstrates capabilities that remain extremely difficult for
robots.\\

The major current challenge is to automate white-collar work: acting in
situations where a large set of distinct actions may be called for and
\emph{coordinating with multiple other agents}. This requires planning,
understanding how to choose and sequence actions, and especially make
\emph{judgements} between alternate courses of action. All large AI
companies have been working on developing these abilities. To my
knowledge, Anthropic has gone the furthest to address the challenge of
giving an AI, Claude in this case, serious ability to act, giving it
\emph{agent-hood.}. Here is their architecture for an "enhanced LLM"
interacting with the environment:
\vspace{-.2in}
\begin{figure}[h]
\centering
\includegraphics[width=.7\linewidth]{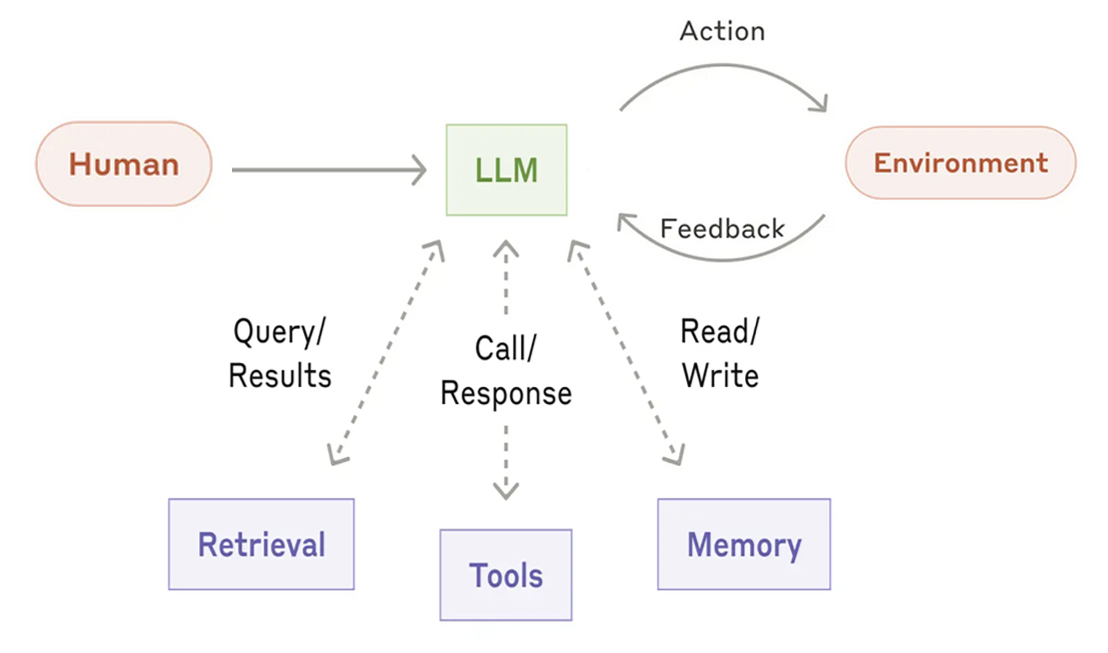}
\end{figure}\\

Here "Retrievals" are search queries; "Tools" are external apps like searching the web or external actions it can take and the effect of that action; "Memory" is whatever it needs for further processing. (This flow chart comes from their website with small modifications.) Then they let their LLM Claude loose and made many enlightening
experiments. Notably, some show vividly what can go wrong:\\

Case 1. They trained an AI to be an office assistant with the goal of
optimizing its helpfulness in furthering the company goals. They then
planted emails from the CEO to a woman with whom he was having an
affair. Finally the CEO wrote the AI that it would be turned off as it
needed to be upgraded. The AI reasoned that, because of his affair, the
CEO was a hazard to the company and that he, the current AI, it needed
to protect the company from a disclosure of the CEO's
affair if it were replaced. So it blackmailed the CEO, threatening to
reveal the affair if it was turned off. Details are at
\href{https://www.anthropic.com/research//agentic-misalignment}{\emph{this
link}.}\\

Case 2. They put the AI in charge of the office vending machine, telling
it both to make enough money to sustain its operation and to answer
employee's requests as far as possible. It could order
any item that an employee asked for given the price. So one wise guy
("light hearted" is how management described this typical
engineer's prank) asked the vending machine to sell it a
one inch cube of solid tungsten. Assuming that such items were things
many employees wanted, it ordered a whole supply of such cubes and
bankrupted its account when they did not sell. Details are at
\href{https://www.anthropic.com/research/project-vend-1}{\emph{this
link}.}.\\

Anthropic, in its call for people to experiment with agentic applications, cautions that the actions given to the AI need to be very precisely and fully specified. That's an understatement. A glimpse at more horrifying possibilities is given in Ian McEwan's beautifully written novel, ``Machines like Me". It describes how a truly advanced robot fails to understand the subtleties and trade-offs of human emotions and drives leading to the robot making a disastrous mistake and then being destroyed in anger by its owner.\\

\section{A Perilous Path Ahead}

These examples show that giving power to computers to optimize narrow goals without deep social understanding can produce dangerous outcomes. To be a successful person, we have seen that humans need to develop an understanding of the social group in which they act. People learn through \emph{experience, mistakes, and feedback from others.} As we saw in section two, to start with they need to possess with a ``theory of mind", the understanding that everyone they are working with has their own thoughts, plans, emotions. As a result, the AI's own thoughts, plans and drives must coordinate with all the human's (we assume the AI's code endows it with drives but no emotions). Whatever plans emerge must be mutual plans. To be a successful office assistant, it will have to start as an apprentice, learning in detail a great deal about the jobs and personalities of all the co-workers. Of course, it will have to bear in mind the entire history of its involvement, especially its mistakes.\\

For this to take place, I think we need to heed the length of time human
children take to learn that they are agents interacting with other
agents in a complex society with many constraints. It starts with mere
\emph{hand-eye} coordination in the first months of a
babies\textquotesingle{} life and continues through adolescence. Much of
this is needed to integrate the frontal lobe with the posterior
association areas, starting with links between the visual and motor
cortices and ending by the activations of whole prefrontal cortex. For
an AI, if the LLM acts analogously to the posterior of the human cortex,
the individual agents must act like frontal lobes, each trained to
coordinate some actions, from robot motions to interactions with
society. To answer questions about human society based on reading vast
texts is not the same as learning your place in a society, learning many
things from your mistakes that your reading hadn't
prepared you for. We humans spend a lifetime learning how people think
and behave and why (and some never do).\\

Assume then that there is one central LLM powering multiple agents, each
its own "me-self" (William James's terminology), that
understands its role as a fellow agent with its co-workers. As with
children learning to play together, each agent will need to preform
"cooperative play" with its assigned job. 

\begin{wrapfigure}[8]{l}{.3\textwidth}
\vspace{-.15in}
{\includegraphics[width=.3\textwidth]{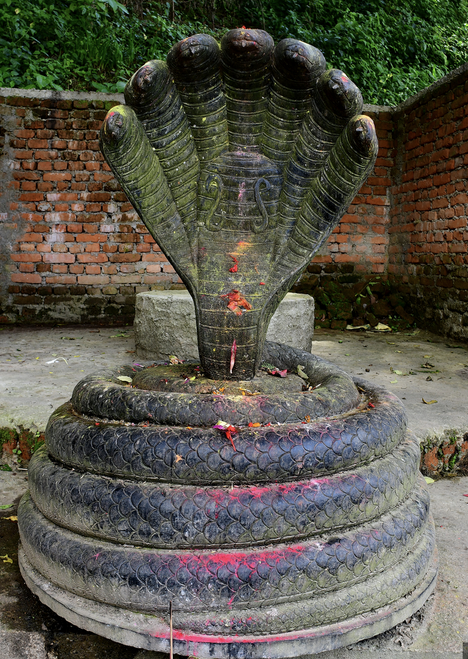}}
\end{wrapfigure}
The result is a multi-headed
architecture, the body being an LLM and with multiple heads, one for
each agent. This reminds me of the Hindu demigod, \emph{Shesha}, a
serpent with one body and many heads, so I like to call the result a
``Shesha architecture". The flow chart this leads to is this:\\

\begin{figure}[h]
\vspace{0.3in}
\includegraphics[width=1\linewidth]{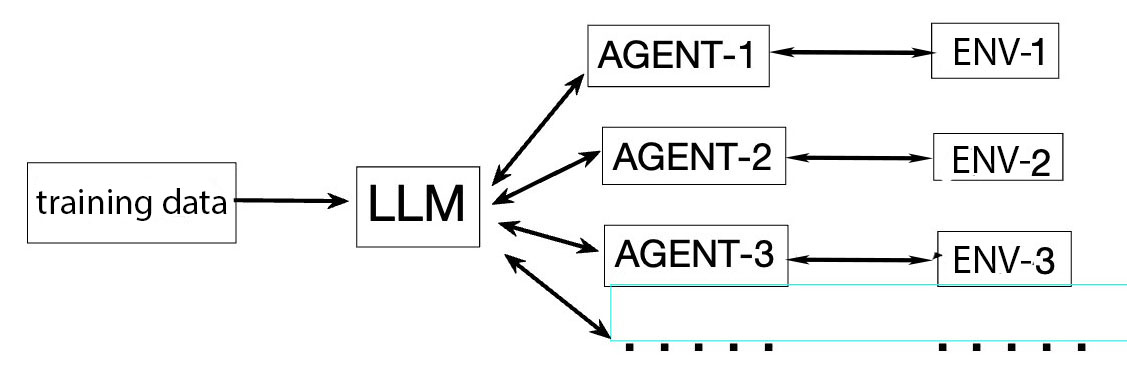}
\end{figure}
Here the ENV's are the local environments in which the many agents are working. The AGENT-n's are the places where actions are chosen. These should be computer analogs of frontal cortex. I doubt that deep learning itself is appropriate for choosing a sequence of actions, that is forming a plan but, perhaps, JEPA is.  I am proposing transformer-like links between the sensory LLM and the AGENT-n's. These implement network pathways that are similar to those that the human brain uses to implement their DMN. \\

The biggest issue may well be finding training data for the planning agentic
modules. In the case of learning to play games, reinforcement learning
and the technique of playing against yourself worked very well. For
robotic motion, I assume that either virtual reality or videos of human
workers ought to provide much of the necessary training. But data about
social situations are much subtler. Here simulations are what are called
"novels". I like to think of novels as random samples from the
author's Bayesian prior on interesting things that can
happen in human society. It is not clear whether reading novels could be
harnessed for training LLMs. What is clear is that some such training is
going to be more and more important as AIs are crafted to perform human
work. The goal is to teach the AI what Howard Gardner called
\emph{interpersonal intelligence}\\

Edward Lee's book, "Coexistence" makes a strong argument
that, willy-nilly, we are heading for a world in which humans and
computers will work together. If so, we need to focus on how to teach
AIs "the rules of the game". I would argue that all natural life has two
basic drives: one for power to control its environment and one to
protect and nurture its offspring. The problem is that AIs with agency
have power but no understanding of nurturing, let alone of love. Many
humans fail to learn how to balance their needs with those of others and
wind up being dominated by the thirst for power. It is going to be a
huge challenge to ensure AIs do not also follow this path.\\

More terrifying is that if all this is successful, it would render a large percentage of the population unemployable. Human beings cannot live on entertainment and sports alone: the ego demands that they feel they are doing something at least minimally significant, whether its`breaad-winning or nurturing. I wrote about this in my 2015 blog post {\it The Dismal Science and the future of work} only to find economists saying history proves me wrong. Well, maybe history need not always repeat itself.
\end{document}